\pdfoutput=1

\documentclass[11pt,dvipsnames]{article}

\usepackage[]{acl}

\usepackage{times}
\usepackage{latexsym}

\usepackage[T1]{fontenc}

\usepackage[utf8]{inputenc}

\usepackage{microtype}

\usepackage[utf8]{inputenc} 
\usepackage[T1]{fontenc}    

\usepackage{url}            
\usepackage{booktabs}       
\usepackage{amsfonts}       
\usepackage{nicefrac}       
\usepackage{microtype}      
\usepackage{xcolor}         
\usepackage{multirow}
\usepackage{float}
\usepackage{amssymb}
\usepackage{amsmath}
\usepackage{bm}
\usepackage{latexsym}
\usepackage{subfigure}
\usepackage{graphicx}
\usepackage[ruled,vlined]{algorithm2e}
\usepackage{color, soul, colortbl}
\usepackage{xcolor}
\definecolor{myred}{rgb}{0.839,0.341,0.271}
\definecolor{myblue}{rgb}{0.843,0.898,0.941}
\usepackage{tikz}
\usepackage{makecell}
\usetikzlibrary{positioning}
\usepackage{cleveref}
\usepackage{ragged2e}

\crefname{section}{§}{§§}
\Crefname{section}{§}{§§}

\SetCommentSty{mycommfont}

\title{Few-shot Classification with Hypersphere Modeling of Prototypes}

\author{Ning Ding\thanks{\ \  Equal contribution} $\ ^1$, Yulin Chen$^{*1}$, Ganqu Cui$^1$,  {Xiaobin Wang}$^2$ \\ \textbf{Hai-Tao Zheng}$^1$, \textbf{Zhiyuan Liu}$^1$, \textbf{Pengjun Xie}$^2$ \\ 
  $^1$Tsinghua University, $^2$Alibaba Group \\
\texttt{\{dingn18, yl-chen21\}@mails.tsinghua.edu.cn}
  }

\begin{document}
\maketitle
\begin{abstract}
\looseness=-1 
Metric-based meta-learning is one of the de facto standards in few-shot learning. 
It composes of representation learning and metrics calculation designs. 
Previous works construct class representations in different ways, varying from mean output embedding to covariance and distributions.
However, using embeddings in space lacks expressivity and cannot capture class information robustly, while statistical complex modeling poses difficulty to metric designs.
In this work, we use tensor fields (``areas'') to model classes from the geometrical perspective for few-shot learning. 
We present a simple and effective method, dubbed as hypersphere prototypes (\texttt{HyperProto}),  where class information is represented by hyperspheres with dynamic sizes with two sets of learnable parameters: the hypersphere's center and the radius.  
Extending from points to areas, hyperspheres are much more expressive than embeddings. Moreover, it is more convenient to perform metric-based classification with hypersphere prototypes than statistical modeling, as we only need to calculate the distance from a data point to the surface of the hypersphere. Following this idea, we also develop two variants of prototypes under other measurements. 
Extensive experiments and analysis on few-shot learning tasks across NLP and CV and comparison with 20+ competitive baselines demonstrate the effectiveness of our approach.

\end{abstract}

\section{Introduction}
\label{sec:intro}

\looseness=-1 Learning from a few examples, i.e., few-shot learning, is receiving increasing amounts of attention in modern deep learning. Because constituting cognition of novel concepts with few instances is crucial for machines to imitate human intelligence, and meanwhile, annotating large-scale supervised datasets is expensive and time-consuming~\citep{lu2020learning}. 
Although traditional deep neural models have achieved tremendous success under sufficient supervision, it is still challenging to produce comparable performance when training examples are limited. 
Hence, a series of studies are proposed to generalize deep neural networks to low-data scenarios. One crucial branch of them is metric-based meta-learning~\citep{reed1972pattern, nosofsky1986attention, snell2017prototypical}, where models are trained to generate expressive representations and carry out classification via defined metrics. 

The success of metric-based learning depends on both \textit{representation} learning and the \textit{metrics} chosen.
One straightforward approach relies on training feature representation and adopts a nearest-neighbor classifier \citep{vinyals2016matching, yang-katiyar-2020-simple, wang2019simpleshot}. Other works introduce additional parameters as class representation to achieve better generalization ability. A naive way to estimate class representation is to use the mean embedding of feature representation \citep{snell2017prototypical, allen2019infinite}, while some also use second-order moments~\citep{li2019CovaMNet} or reparameterize the learning process to generate class representation in a richer semantic space~\citep{avinash2019embedclass} or in the form of probability distribution~\citep{zhang2019variational}. 
Apart from traditional Euclidean and cosine distance, a variety of metric functions are also proposed~\citep{sung2018learning, zhang2020deepemd, xie2022joint}. 
Most existing works learn class representation from the statistical perspective, making designing and implementing the metrics more difficult. For example, the proposed covariance metric in CovaMNet~\citep{li2019CovaMNet} theoretically requires a non-singular covariance matrix, which is awkward for neural-based feature extraction methods.

\looseness=-1 This paper revisits metric-based learning and finds that geometrical modeling can simultaneously enhance the expressive ability of representations and reduce the difficulty of calculation, meanwhile yielding surprisingly effective performance in few-shot learning. 
Specifically, we propose \texttt{HyperProto}, a simple and effective approach to model class representation with hyperspheres.
It is equipped with two advantages: the modeling is straightforward, and the corresponding metrics are easier to define and calculate compared to statistical methods.
(1) For one thing, even if we attempt to use geometrical ``areas'' instead of ``points'' to represent class-level information, it is still difficult to explicitly characterize manifolds with complex boundaries in deep learning. But via hyperspheres modeling, we can obtain a hypersphere prototype with only two sets of parameters: the center and the radius of hyperspheres. (2) Besides, hyperspheres are suitable for constructing measurements in Euclidean space. We can calculate the Euclidean distance from one feature point to the surface of the hypersphere in order to perform metric-based classification, which is difficult for other manifolds. 

\looseness=-1 We set the radii of the hyperspheres as learnable parameters, which makes it easy to combine the two advantages in few-shot learning.
The distance from one feature point to the surface of a hypersphere prototype can be formalized as the distance from the point to the center of the hypersphere minus the radius. 
Thus, both the radius and the center of the hypersphere can appear in the loss function and participate in the backward propagation during optimization. Intuitively, for the classes with sparse feature distributions, the corresponding radii of their prototypes are large, and the radii are small otherwise.
Beyond the Euclidean space, we also develop two variants based on the general idea -- cone-like prototypes with cosine similarities and Gaussian prototypes from the probability perspective. 

\looseness=-1 We conduct extensive experiments to evaluate the effectiveness of \texttt{HyperProto}.  In addition to two classical tasks, few-shot named entity recognition (NER)~\citep{ding2021fewnerd} and relation extraction (RE)~\citep{han2018fewrel, gao2019fewrel} in NLP, we also assess our approach on few-shot image classification~\citep{vinyals2016matching, welinder2010caltech}, proving that it is a general method that could be applied to diverse scenarios.
 Despite the simplicity, we find that our approach is exceedingly effective, which outperforms the vanilla prototypes by 8.33 \% absolute in average F1 on \textsc{Few-NERD (INTRA)}, 6.55\% absolute in average F1 on \textsc{Few-NERD (INTER)}, 4.77\% absolute in average accuracy on FewRel, 21.63\% absolute in average accuracy on FewRel 2.0, and 3.45\% absolute in average accuracy on \textit{mini}ImageNet, respectively. 
Our method also yields better performance with 20+ competitive approaches across three tasks. 
Surprisingly, \texttt{HyperProto} performs more than satisfactorily in cross-domain few-shot relation extraction and cross-dataset image classification, indicating the promising ability in domain adaptation. 
Given that such small changes can bring considerable benefits, we believe our approach could serve as a strong baseline for few-shot learning and inspire new ideas from the research community for representation learning. 

\section{Problem Setup}
\label{subsec:problem}

\looseness=-1 We consider the episodic $N$-way $K$-shot few-shot classification paradigm\footnote{
For the few-shot named entity recognition task (sequence labeling), 
the sampling strategy is slightly different (details in Appendix~\ref{appendix:sample}).}. Given a large-scale annotated training set $\mathcal{D}_{\text{train}}$, our goal is to learn a model that can make accurate predictions for a set of new classes $\mathcal{D}_{\text{test}}$, containing only a few labeled examples for training. The model will be trained on episodes constructed using $\mathcal{D}_{\text{train}}$ and tested on episodes based on $\mathcal{D}_{\text{test}}$. Each episode contains a \textit{support} set  $\mathcal{S} = \{\bm{x}_i, {y}_i \}_{i=1}^{N \times K}$ for learning, with $N$ classes and $K$ examples for each class, and a \textit{query} set for inference $\mathcal{Q} = \{\bm{x}_j^*, {y}_j^* \}_{j=1}^{N \times K'}$ of examples in the same $N$ classes. Each input data is a vector $\bm{x}_i \in \mathbb{R}^L$ with the dimension of $L$ and $y_i$ is an index of the class label. For each input $\bm{x}_i$, let $f_\phi(\bm{x}_i) \in \mathbb{R}^D$ denote the $D$-dimensional output embedding of a neural network $f_\phi: \mathbb{R}^L \rightarrow \mathbb{R}^D$ parameterized by $\phi$.

\section{Methodology}
\label{sec:bp}

\looseness=-1 This section introduces the mechanisms of hypersphere modeling of prototypes. 
One hypersphere prototype is represented by two parameters: the center and the radius of the hypersphere, which is first initialized via estimation and then optimized by gradient descent along with the encoder parameters.



\begin{figure*}[!th]
 	\centering
 		\includegraphics[width=0.93\linewidth]{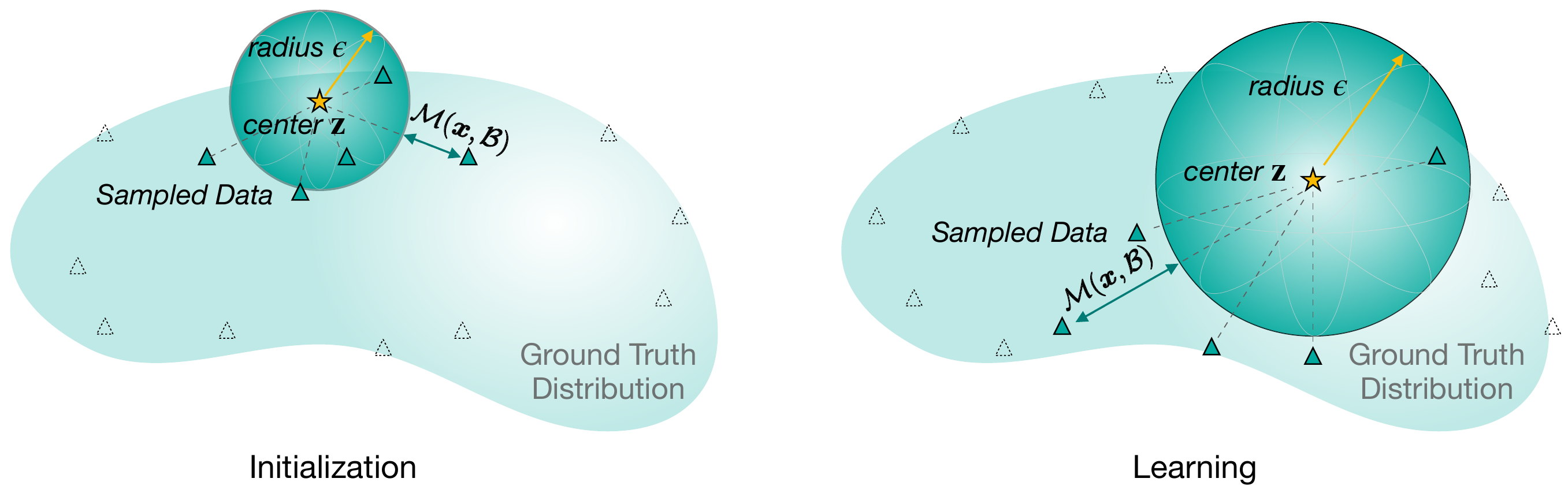}
 		\caption{The illustration of our proposed \textit{HyperProto}, where the data is sampled in 5-shot. The star symbol denotes the center of the hypersphere, the solid triangle denotes the sampled examples, and the dotted triangle denotes other examples in the whole dataset. The solid green line denotes the distance from a data embedding to the hypersphere's surface. The left part illustrates the initialization stage, where the sampled data estimate the center and radius, and the right part illustrates the learning stage, where the center and radius are simultaneously optimized. 
 		}
 		
\label{fig:tsne-intro}
\end{figure*}

\subsection{Overview}
\label{subsec:model}

We now introduce \texttt{HyperProto}, which are a set of hyperspheres in the embedding space $D$ to abstractly represent the intrinsic features of classes. Formally, one prototype is defined by
\begin{equation}
  \mathcal{B}^d(f_\phi, \bm{z}, \epsilon):=\{f_\phi(\bm{x})\in \mathbb{R}^D: d(f_\phi(\bm{x}), \bm{z}) \leq \epsilon\},  
\end{equation}
where $d: \mathbb{R}^D \times \mathbb{R}^D \rightarrow [0, +\infty)$ is the distance function in the metric space.  
$f_\phi$ is a neural encoder parameterized by $\phi$, while $z$ and $\epsilon$ denote the center and the radius of the hypersphere. We use $\mathcal{M}(\cdot)$ to denote the measurement between a data point and a hypersphere prototype based on $d(\cdot)$.


The central idea is to learn a hypersphere prototype for each class with limited episodic supervision, and each example in the query set $(\bm{x}^*, y^*)$ is predicted by the measurement to the hypersphere prototypes $\mathcal{M}(\bm{x}^*_j, \mathcal{B}^d)
$, which is the Euclidean distance from the embedding to the \textit{surface} of the hyperspheres,
\begin{equation}
\label{eq:metric}
    {\mathcal{M}}(\bm{x}, \mathcal{B}) = d(f_\phi(\bm{x}), \bm{z}) - \epsilon = \left \|f_\phi(\bm{x}) - \bm{z}  \right\| ^2_2 - \epsilon.
\end{equation}
\looseness=-1 Note that in this case, the value of $\mathcal{M}(\cdot)$ may be negative. That is, geometrically speaking, the point is contained inside the hypersphere, which does not affect the calculation of the loss function and the prediction. 
Generally, the idea is to use areas instead of points in the embedding space to model prototypes, and hyperspheres naturally have two advantages. First, as stated in \cref{sec:intro}, one hypersphere prototype could be uniquely modeled by the center $\bm{z}$ and the radius $\epsilon$, while characterizing manifolds with complex boundaries in the embedding space is intricate. Second, it is easy to optimize the parameters by conducting metric-based classification since they are naturally involved in measurement calculation.
In this geometric interpretation, sparse classes will have larger learned radii, while compact classes will have smaller learned radii.


\subsection{Hypersphere Prototypes}
\label{subsec:init}

\looseness=-1 To construct hypersphere prototypes, the first step is the initialization of the center $\bm{z}$ and the radius $\epsilon$ of the hypersphere. To start with a reasonable approximation of the data distribution, we randomly select $K$ instances from each class for initialization. Specifically for one class, the center of the hypersphere prototype is the mean output of the $K$ embeddings as the estimation in~\citet{snell2017prototypical}, and the radius is the mean of the distance of each sample to the center.  $\mathcal{S}_n$ 
is the set of sampled instances from the $n$-th class,
\begin{equation}
\label{eq:init}
\mathcal{B}_n:=\left\{
\begin{aligned}
\bm{z}_n &= \frac{1}{K}\sum\limits_{\bm{x}\in\mathcal{S}_n} f_\phi(\bm{x}), \\
\epsilon_n &= \frac{1}{K} \sum\limits_{\bm{x}\in\mathcal{S}_n} d(f_\phi(\bm{x}),\bm{z}_n).
\end{aligned}
\right.
\end{equation}
Once initialized, a hypersphere prototype will participate in the training process, where its center and radius are simultaneously optimized. 
During training, for each episode, assume the sampled classes are $\mathcal{N}=\{n_1, n_2, ..., n_N\}$, the probability of one query point $\bm{x} \in \mathcal{Q}$ belonging to class $n$ is calculated by softmax over the metrics to the corresponding $N$ hypersphere prototypes.
\begin{equation}
\label{eq:p}
\begin{split}
    p(y = n|\bm{x}; \phi) = \frac{\exp (-\mathcal{M}(\bm{x}, \mathcal{B}_n))}{\sum_{n' \in \mathcal{N}} \exp (-\mathcal{M}(\bm{x}, \mathcal{B}_{n'}))}.
\end{split}
\end{equation}
And the parameters of $f$ and hypersphere prototypes are optimized by minimizing the  metric-based cross-entropy objective:
\begin{equation}
\label{eq:cls}
    \mathcal{L}_\text{cls} = - \log p(y|\bm{x}, \phi, \bm{z}, \epsilon).
\end{equation}

\looseness=-1 Equation~\ref{eq:p} explains the combination of the advantages of hypersphere prototypes, where $\mathcal{M}$ is calculated by $r$ and $\bm{z}$, which will participate in the optimization. The parameters of the neural network $\phi$ are optimized along with the centers and radii of hypersphere prototypes through gradient descent. 
To sum up, in the initialization stage, the hypersphere prototypes of all classes in the training set, which are parameterized by $\bm{z}$ and $\epsilon$, are estimated by the embeddings of randomly selected instances and \textit{stored} for subsequent training and optimization. 
In the learning stage, the stored $\epsilon$ is optimized by an independent optimizer, because, empirically, the parameter could benefit from large learning rates. The optimization will yield a final location and size of the hyperspheres to serve the classification performance. More importantly, the involvement of  prototype centers and radii in the training process will, in turn, affect the optimization of encoder parameters, stimulating more expressive and distinguishable representations.

Algorithm~\ref{alg:training} expresses the initialization and learning stages of hypersphere prototypes. 
Although the centers and radii are stored and optimized continuously in training (in contrast with vanilla prototypes where centers are re-estimated at each episode), the whole process is still largely episodic, as in each episode, the samples in the query set are only evaluated against the classes in that single episode instead of the global training class set.

{

\begin{algorithm*}[h]
\caption{\looseness=-1 Training process. $f_{\phi}$ is the feature encoder, 
$N_\text{total}$ is the total number of classes in the training set, $N$ is the number of classes for support and query set, $K$ is the number of examples per class in the support set, $K'$ is the number of examples per class in the query set, 
$M$ is a hyper-parameter. $\textsc{RandomSample}(S, K)$ denotes a set of $K$ elements chosen uniformly at random from set $S$, without replacement. $\lambda_{f}$ and $\lambda_{\epsilon}$ are separate learning rates.}
\label{alg:training}
\KwIn{Training data $\mathcal{D}_\text{train}=\{(\bm{x}_1, y_1),...,(\bm{x}_{T}, y_{T})\}$, $y_i \in \{1,..., N_\text{total}\}$. $\mathcal{D}_k$ denotes the subset of $\mathcal{D}$ containing all elements $(\bm{x}_i, y_i)$ such that $y_i=k$}
\KwOut{The updated encoder $f_{\phi}$}%
{\tcp{Initialization phase}}
\For{$n=1$ to $N_\text{total}$}{ 
  $\mathcal{S}_{n}\leftarrow \textsc{RandomSample}(\mathcal{D}_n, K)$  \\
  $\bm{{z}}_n \leftarrow   \frac{1}{|\mathcal{S}_{n}|} \sum\limits_{(\bm{x}_i, y_i) \in \mathcal{S}_{n}} \!\! f_\phi (\bm{x}_i)$, \\
  ${\epsilon}_n \leftarrow \frac{1}{|\mathcal{S}_{n}|} \sum\limits_{(\bm{x}_i, y_i) \in \mathcal{S}_{n}} d(f_\phi(\bm{x}_i), \bm{z}_n), $
}
{\tcp{Learning phase}}
\For{$i=1$ to $M$}{
    $V \leftarrow  \textsc{RandomSample}(\{1,...,N_\text{total}\}, N)$,  $\mathcal{L}_{\text{cls}} \leftarrow 0$ \\
    \For{$n$ in $\{1,...,N\}$}{
    $\mathcal{Q}_{n}\leftarrow \textsc{RandomSample}(\mathcal{D}_{V_n}, K')$ \\
    
    $\mathcal{L}_{\text{cls}} \leftarrow \mathcal{L}_{\text{cls}} + \frac{1}{NK'}\sum\limits_{(\bm{x}_i, y_i) \in \mathcal{Q}_{n}} [d(f_\phi(\bm{x}_i), \bm{z}_n)-\epsilon_n+\log\sum\limits_{n'}\exp(\epsilon_{n'}-d(f_\phi(\bm{x}_i), \bm{z}_{n'}))]$ \\
}
\textsc{Update} $\bm{z}$, $\epsilon$, $f_{\phi}$ w.r.t $\mathcal{L}_{\text{cls}}$, $\lambda_{f}$, $\lambda_{\epsilon}$
}
\end{algorithm*}
}

\looseness=-1 Meanwhile, a standard episodic evaluation process is adopted to handle the unseen classes, where we estimate prototype centers and radii in closed forms. In the episodic evaluation procedure, \texttt{HyperProto} directly takes the mean of instance embeddings as the centers and the mean distance of each instance to the center as the radius (as in Equation~\ref{eq:init}), following previous standard~\citep{vinyals2016matching, snell2017prototypical, zhang2020deepemd}. 

\subsection{Generalizations}
\label{sec:gen}

We have introduced the mechanisms of hypersphere prototypes in Euclidean space. In this section, we generalize this idea to construct variants with other measurements. 


\noindent \looseness=-1 \textbf{Cone-like Prototypes.} Cosine similarity is a commonly used measurement in machine learning. Assume all the data points are distributed on a unit ball, and we use the cosine of the angle to measure the similarity of the two embeddings. 
While keeping the intuition of the modeling of hypersphere prototypes in mind, we introduce an additional angle parameter $\epsilon$. We use $\theta_{\bm{a}, \bm{b}}$ to denote the angle of the two embeddings $\bm{a}$ and ${\bm{b}}$.
In this way, the center point $\bm{z}$ and the angle $\epsilon$ could conjointly construct a cone-like prototype,
\begin{equation}
    \mathcal{B}^d(\bm{z}, f_\phi, \epsilon)\!\! :=\!\! \{f_\phi(x) \in \mathbb{R}^D\!:\! d(f_\phi(\bm{x}), \bm{z}) \geq \cos\epsilon\},
\end{equation}
where $d(f_\phi(\bm{x}), \bm{z})= \cos(\theta_{f_\phi(\bm{x}), \bm{z}})$.
The measurement $\mathcal{M}(\cdot)$ is defined as the cosine of the angle between the instance embedding and the nearest point on the border of the prototype,
\begin{equation}
\label{eq:cos}
\mathcal{M}(\bm{x}, \mathcal{B}) =\! \left\{
\begin{aligned}
& \!\! -\cos (\theta_{f_\phi(\bm{x}), \bm{z}} -\epsilon), \theta_{f_\phi(\bm{x}), \bm{z}} \geq |\epsilon |, \\
&\!\! -1, \theta_{f_\phi(\bm{x}), \bm{z}} < |\epsilon |.
\end{aligned}
\right.
\end{equation}

\begin{figure}[h]
\centering
\includegraphics[width=1\linewidth]{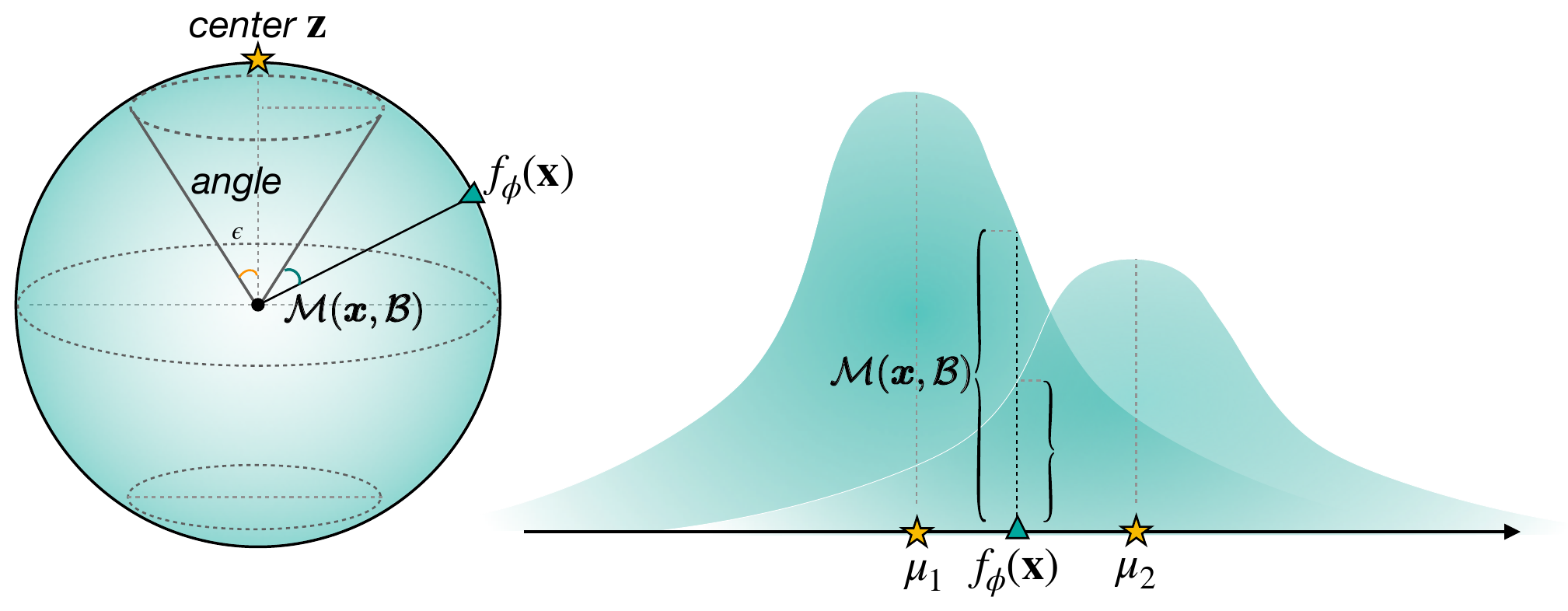}
\caption{Two variants according to different measurements. The left is the cone-like modeling with cosine similarities, and the right is the Gaussian modeling from the probability perspective.}
\label{fig:var}
\end{figure}


\looseness=-1 Similar to the vanilla hypersphere prototypes, $\bm{z}$ and $\epsilon$ need to participate in the learning process for optimization, and the angle $\theta_{\bm{x},\bm{z}}$ is computed by the inverse trigonometric function,
\begin{equation}
    \theta_{f_\phi(\bm{x}),\bm{z}} = \arccos{\frac{f_\phi(\bm{x})^T\bm{z}}{||f_\phi(\bm{x})||\cdot||\bm{z}||}}.
\end{equation}

The prediction for a training example is also based on the softmax over the measurements to the  prototypes like Eq.~\ref{eq:cls}.
Note that as shown in Eq.~\ref{eq:cos}, the measurement becomes $-1$ when a data point is ``inside'' the cone-like  prototype. Then it is hard to make a prediction when an embedding is inside two prototypes. It thus requires that the prototypes do not intersect with each other, that is, to guarantee the angle between two center points is larger than the sum of their own parameter angles,
\begin{equation}
    \mathcal{L}_\text{dis} = \frac{1}{N} \sum_{i, j} \max((|\epsilon_i | + |\epsilon_j |) - \theta_{\bm{z}_i, \bm{z}_j}, 0).
\end{equation}
Therefore, the final loss function is $\mathcal{L} = \mathcal{L}_\text{cls} + \mathcal{L}_\text{dis}$.

\noindent \textbf{Gaussian Prototypes.} From the probability perspective, each class can be characterized by a distribution in a multi-dimensional feature space. The measurement of a query sample to the $n$-th class can thus be represented by the negative log likelihood of $f_\phi(\bm{x})$ belonging to $\mathcal{B}_n$. In line with other works~\citep{zhang2019variational,li2020asymmetric}, we can simply assume each class subjects to a Gaussian distribution $\mathcal{B}_n\sim \mathcal{N}(\bm{\mu}_n, \Sigma_n)$. To reduce the number of parameters and better guarantee the positive semi-definite feature, we can further restrict the covariance matrix to be a diagonal matrix such that $\Sigma_n=\sigma_n^2 I$. Then the measurement becomes
\begin{equation}
\begin{split}
    \mathcal{M}&(\bm{x}, \mathcal{B}_n)\!\! = \!\! -\log p(f_{\phi}(\bm{x});\mathcal{B}_n) 
    \\
    &=\!\!\frac{||f_{\phi}(\bm{x})-\bm{\mu}_n||_2^2}{2\sigma_n^2} +\log((2\pi)^{\frac{d}{2}}|\sigma_n|^{d}) \\
    &=\!\!\frac{||f_{\phi}(\bm{x})-\bm{\mu}_n||_2^2}{2\sigma_n^2} +d\log |\sigma_n|+\delta,
    \label{eq:dis_metric}
\end{split}
\end{equation}
\looseness=-1 where $\delta=\frac{d}{2}\log2\pi$. The probability of target class given a query sample can be calculated by Eq.~\ref{eq:p} in the same fashion: $     p(y=n|\bm{x})=\frac{p(f_{\phi}(\bm{x});\mathcal{B}_n)}{\sum_{n'}p(f_{\phi}(\bm{x});\mathcal{B}_{n'})}
     \label{eq:dis_p}$.
Note that the derived form of the equation is the same as directly calculating the probability of $p(y=n|\bm{x})$ under a uniform prior distribution of $p(y)$. Comparing with pure probabilistic approaches, such as variational inference that treats $\mathcal{B}$ as hidden variables and models $p(\mathcal{B}|\mathcal{S})$ and $p(\mathcal{B}|\mathcal{S}, \bm{x})$ with neural network~\citep{zhang2019variational}, under the framework of \cref{subsec:init}, $\theta$ is explicitly parameterized and optimized for each class during training. Moreover, comparing Eq.~\ref{eq:dis_metric} with Eq.~\ref{eq:metric}, it can be observed that when formalizing $\mathcal{B}$ as a distribution, instead of as a bias term, the original radius parameter (now the variance) functions as a scaling factor on Euclidean distance.

\section{Experiments}
\label{sec:exp}

\looseness=-1 To evaluate the effectiveness of the proposed method, we conduct experiments on three few-shot learning tasks in NLP and CV, including few-shot named entity recognition (NER), few-shot relation extraction (RE), and few-shot image classification. We chose these three tasks because they all have well-established datasets and baselines to facilitate comprehensive comparisons, while they are still challenging under the few-shot setting as fundamental tasks in NLP and CV. 
Apart from the experimental study in this section, additional experiments and analyses are reported in Appendix~\ref{appendix:additional_exp}. The task descriptions, datasets, and implementation details are reported in Appendix~\ref{appendx:exp}. Techniques like the structures of neural models, task-specific pre-training, and distillation are \textit{orthogonal} to our contributions. 

\subsection{Comparison with Embedding Prototypes}

\begin{figure*}[!htb]
    \centering
    \includegraphics[width=0.95\textwidth]{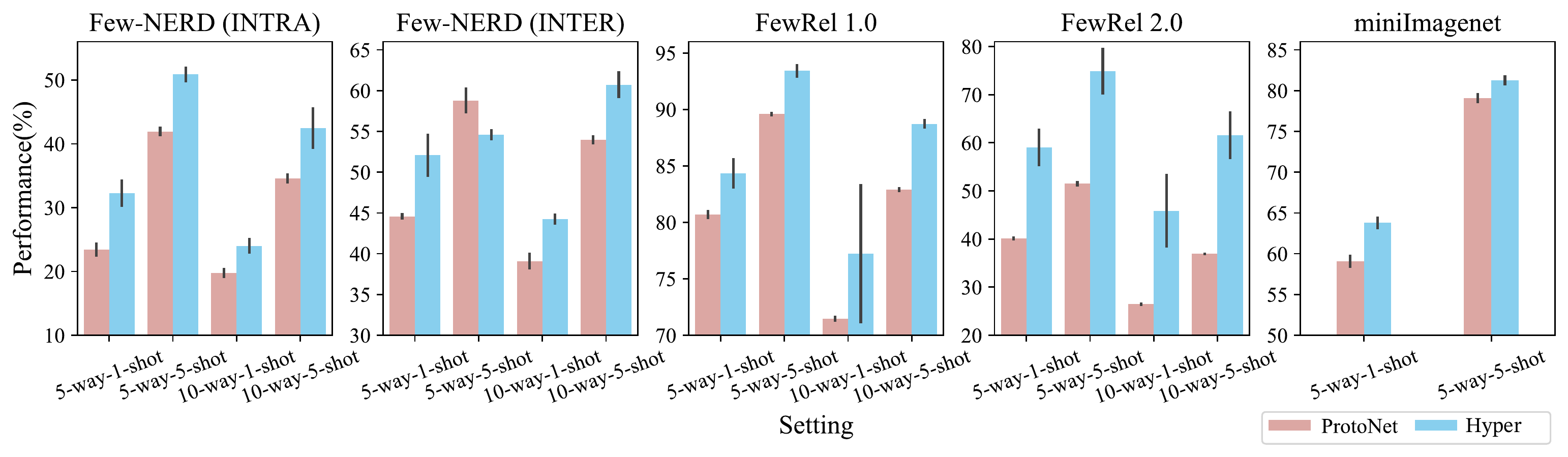}
    \caption{Performance comparison between vanilla prototypes and our hypersphere prototypes, we use the same BERT model as the encoder for all experiments in named entity recognition and relation extraction.}
    \label{fig:p_bigp}
\end{figure*}

We first investigate the direct effect of the proposed geometrical modeling on few-shot learning.
We choose the vanilla modeling of prototypes, i.e., using vectors with the same dimension of hidden representations, as our primary baseline. 
As illustrated in Figure~\ref{fig:p_bigp}, our method consistently outperforms vanilla prototypes on all the tasks. Specifically for named entity recognition, our method yield an increase of at least 5\% in f1-score across all settings; for relation extraction, our method achieves improvements of 3\% $\sim$ 19\% in f1-score across all settings; for image classification, the improvements are range from 2\% $\sim$ 4\%. Since all other settings in this experiment are identical, the results could directly demonstrate the superiority of hypersphere modeling.

\newcolumntype{g}{>{\columncolor{myblue}}c}
\begin{table*}[!ht]

\centering
\scalebox{0.88}{
\begin{tabular}{l|cccgccg}
\toprule
\multirow{2}{*}{Setting}                                                          & \multirow{2}{*}{Eva.} & \multicolumn{3}{c}{\textsc{Few-NERD (INTRA)}} & \multicolumn{3}{c}{\textsc{Few-NERD (INTER)}} \\ \cmidrule{3-8}
                                                                                  &                         & NNShot   & ProtoNet    & \texttt{HyperProto}  & NNShot   & ProtoNet    & \texttt{HyperProto}  \\ 
                                        \midrule
\multirow{3}{*}{\begin{tabular}[c]{@{}l@{}}5 way\\  1$\sim$2 shot\end{tabular}}    & P          &  28.95 ± $\text{1.02}$      &      18.58 ± $\text{1.02}$ &       {40.18} ± $\text{1.71}$       &  50.40 ± $\text{0.60}$  
     &  38.70 ± $\text{0.50}$  &       {53.36} ± $\text{2.74}$      \\ 
                                                                                  & R                       & {33.40} ± $\text{1.44}$       &     31.83 ± $\text{1.03}$  &         26.96 ± $\text{2.07}$     &  {58.84} ± $\text{0.13}$      &   52.60 ± $\text{1.65}$ &     51.12 ± $\text{4.94}$       \\ 
                                                                                  & F                      &   31.01 ± $\text{1.21}$     &  23.45 ± $\text{0.92}$    &        {32.26} ± $\text{1.94}$       &    {54.29} ± $\text{0.40}$   & 44.58 ± $\text{0.26}$   &          52.09 ± $\text{2.49}$      \\
                                        \midrule
\multirow{3}{*}{\begin{tabular}[c]{@{}l@{}}5 way\\ 5$\sim$10 shot\end{tabular}}   & P                       &    32.87 ± $\text{2.45}$      &  35.87 ± $\text{0.69}$   &             {48.77 ± $\text{0.79}$}      &  45.80 ± $\text{3.53}$        &  53.73 ± $\text{1.77}$         &      {62.26 ± 0.89}      \\
                                                                                  & R                       &   39.17 ± $\text{2.17}$     &    {50.50 ± $\text{1.88}$}  &        {53.26 ± 2.60}        &  56.45 ± $\text{2.93}$      & 64.99 ± $\text{2.24}$     &      {69.32 ± 1.66}       \\
                                                                                  & F                      & 35.74 ± $\text{2.36}$      &    41.93 ± $\text{0.55}$   &      {50.88 ± 1.01}        &  50.56 ± $\text{3.33}$       &  58.80 ± $\text{1.42}$   &          {65.59 ± 0.50}      \\ 
                                        \midrule
\multirow{3}{*}{\begin{tabular}[c]{@{}l@{}}10 way\\ 1$\sim$2 shot\end{tabular}}   & P                       &     20.38 ± $\text{0.22}$     & 16.52 ± $\text{0.52}$    &         {26.06} ± $\text{2.40}$       &     42.74 ± $\text{2.05}$     & 32.59 ± $\text{0.22}$   &         {45.38} ± $\text{0.49}$       \\
                                                                                  & R                       &     23.63 ± $\text{0.53}$    &    {24.60} ± $\text{0.72}$  &       22.32 ± $\text{0.54}$        &  {52.16} ± $\text{1.76}$      &  48.91 ± $\text{2.94}$   &         43.22 ± $\text{1.33}$       \\
                                                                                  & F                      &   21.88 ± $\text{0.23}$       &   19.76 ± $\text{0.59}$   &        {24.02} ± $\text{1.06}$   &   {46.98} ± $\text{1.96}$    &    39.09 ± $\text{0.87}$    &          44.26 ± $\text{0.53}$         \\ 
                                        \midrule
\multirow{3}{*}{\begin{tabular}[c]{@{}l@{}}10 way\\ 5$\sim$10 shot\end{tabular}} & P                       &  25.46 ± $\text{0.63}$        & 28.93 ± $\text{0.82}$   &          {38.94} ± $\text{3.39}$      &  45.15 ± $\text{0.81}$       &    47.93 ± $\text{0.45}$ &           {56.38} ± $\text{1.79}$      \\
                                                                                  & R                       &   30.32 ± $\text{1.71}$      &    43.08 ± $\text{0.84}$ &        {46.71} ± $\text{2.48}$      &   56.05 ± $\text{0.37}$     &     61.79 ± $\text{1.73}$  &        {65.84} ± $\text{1.61}$     \\
                    & F                      &  27.67 ± $\text{1.06}$      &   34.61 ± $\text{0.59}$   &       {42.46} ± $\text{3.04}$        &    50.00 ± $\text{0.36}$    &    53.97 ± $\text{0.38}$  &       {60.73} ± $\text{1.47}$    \\ 
                    \midrule
                    Average & F & 29.08 & 29.94 & \textbf{37.41} & 50.46 & 49.11 &  \textbf{55.66} \\
                    \bottomrule    
\end{tabular} }
\caption{Performance on the \textsc{Few-NERD} dataset. P denotes precision, R denotes recall, and F refers to the F1 score. The standard deviation is reported with 3 runs with different random seeds for each model.} 
\label{tab:fewnerd}
\end{table*}

\begin{table*}[!th]
\centering

\scalebox{0.8}{
\begin{tabular}{lccccc} \toprule
\multirow{2}{*}{\textbf{ {Model}}} & \multicolumn{5}{c}{\textbf{ FewRel 1.0}}                                               \\ \cmidrule(r){2-6}
                        & \textbf{5 way 1 shot} & \textbf{5 way 5 shot} &
                        \textbf{10 way 1 shot}& \textbf{10 way 5 shot} & \textbf{Average}\\
                    \midrule          
Meta Net~\citep{munkhdalai2017meta} & 64.46 ± 0.54 & 80.57 ± 0.48 & 53.96 ± 0.56 & 69.23 ± 0.52 
 & 67.06\\
SNAIL~\citep{mishra2017simple} & 67.29 ± 0.26 & 79.40 ± 0.22 & 53.28 ± 0.27 & 68.33 ± 0.26 & 67.08 \\
GNN CNN~\citep{satorras2018few} & 66.23 ± 0.75 &  81.28 ± 0.62 &  46.27 ± 0.80 &  64.02 ± 0.77 & 64.45\\
GNN BERT~\citep{satorras2018few} & 75.66 ± 0.00 &  89.06 ± 0.00&  70.08 ± 0.00 &  76.93± 0.00 & 77.93\\
Proto-HATT~\citep{gao2019hybrid} & 76.30 ± 0.06 & 90.12 ± 0.04 & 64.13 ± 0.03 & 83.05 ± 0.05 & 78.40\\
MLMAN~\citep{ye2019multi} & 82.98 ± 0.20 & 92.66 ± 0.09 & 73.59 ± 0.26 & 87.29 ± 0.15 & 84.13\\ 
MTB~\citep{soares2019matching} & 89.80 & 93.59 & 83.37 & 88.64  & 88.85  \\
REGRAB~\citep{qu2020few} & 90.30 & 94.25 & 84.09 & 89.93 & 89.64\\

\midrule
 ProtoCNN   &   {69.20 ± 0.20}    &   84.79 ± 0.16    & {56.44 ± 0.22}   &   75.55 ± 0.19 & 71.50   \\ \rowcolor{myblue}
\texttt{HyperProto} CNN (Ours) &  66.05 ± 3.44    &  {87.31 ± 0.93}    & 56.74 ± 1.06  &   {77.87 ± 2.60} & 71.99  \\
\midrule
 ProtoBERT    &  80.68 ± 0.28    &   89.60 ± 0.09   &   71.48 ± 0.15   &   82.89 ± 0.11  & 81.16  \\ 
\rowcolor{myblue}
\texttt{HyperProto} BERT (Ours) &  {84.34 ± 1.23}    &  {93.42 ± 0.50} &     {77.24 ± 6.05}  &   {88.71 ± 0.31}  & 85.93\\
\rowcolor{myblue}
\texttt{HyperProto} BERT+Pretrain (Ours) &   \textbf{89.90 ± 0.43}  & \textbf{96.23 ± 0.21}  &  \textbf{83.55 ± 0.59}  &   \textbf{92.90 ± 0.19} & \textbf{90.65}\\
\midrule

 & \multicolumn{5}{c}{\textbf{FewRel 2.0 Domain Adaptation}} \\ \midrule
 Proto-ADV CNN~\citep{wang2018adversarial} & 42.21 ± 0.09 & 58.71 ± 0.06 & 28.91 ± 0.10 & 44.35 ± 0.09 & 43.55\\
 Proto-ADV BERT~\citep{gao2019fewrel} & 41.90 ± 0.44 & 54.74 ± 0.22 & 27.36 ± 0.50 & 37.40 ± 0.36 & 40.35\\
 BERT-pair~\citep{gao2019fewrel} & 56.25 ± 0.40 & 67.44 ± 0.54 & 43.64 ± 0.46 & 53.17 ± 0.09 & 55.13\\ \midrule
 ProtoCNN    &   35.09 ± 0.10    &   49.37 ± 0.10    & {22.98 ± 0.05}   &   35.22 ± 0.06    & 35.67\\ 
\rowcolor{myblue}
\texttt{HyperProto} CNN (Ours) &   {36.41 ± 1.43}   &  {55.50 ± 1.42}  &  22.11 ± 0.58   &  {40.82 ± 2.50}  & 38.71\\
\midrule
 ProtoBERT   &   40.12 ± 0.19    &   51.50 ± 0.29   &   26.45 ± 0.10   &  36.93 ± 0.01  & 38.75 \\ 
\rowcolor{myblue}
\texttt{HyperProto} BERT (Ours) &  {59.03 ± 3.68}    &  {74.85 ± 4.59}   &   {45.88 ± 7.43} &   {61.61 ± 4.69}  & 60.34\\
\rowcolor{myblue}
\texttt{HyperProto} BERT+Pretrain (Ours) &  \textbf{73.98 ± 1.72}    & \textbf{90.06 ± 0.45} & \textbf{61.42 ± 0.79}   &  \textbf{81.39 ± 1.25} & \textbf{76.71}\\
\bottomrule
\end{tabular}}

\caption{Accuracies on FewRel 1.0 and FewRel 2.0 under 4 different settings. The standard deviation is reported with 3 runs with different random seeds for each model.} 
\label{tab:fewrel1.0}
\end{table*}

\begin{table*}[!ht]
\centering

\scalebox{0.9}{
\begin{tabular}{lccc} \toprule
\multirow{2}{*}{\textbf{ {Model}}} & \multirow{2}{*}{\textbf{ {Backbone}}} & \multicolumn{2}{c}{\textbf{\textit{mini}ImageNet}}                                               \\ \cmidrule(r){3-4}
                     & &  \multicolumn{1}{c}{ {\textbf{5 way 1 shot}}}  & \multicolumn{1}{c}{ {\textbf{5 way 5 shot}}}  \\
                    \midrule      

Infinite Mixture Prototypes~\citep{allen2019infinite} & ConvNet  & 33.30 ± 0.71 &  65.88 ± 0.71      \\ 

ProtoNet~\citep{snell2017prototypical} & ConvNet  & 46.44 ± 0.60 &  63.72 ± 0.55   \\ 
CovaMNet~\citep{li2019CovaMNet} & ConvNet & 51.83 ± 0.64 & 65.65 ± 0.77 \\
\rowcolor{myblue}
\texttt{HyperProto} (Ours) & ConvNet  & 50.21 ± 0.31 & {66.48 ± 0.71}   \\ \midrule
SNAIL~\citep{mishra2017simple}  & ResNet-12 & 55.71 ± 0.99 & 68.88 ± 0.92 \\
ProtoNet~\citep{snell2017prototypical} & ResNet-12 & 53.81 ± 0.23 & 75.68 ± 0.17  \\
Variational FSL~\citep{zhang2019variational} & ResNet-12 & 61.23 ± 0.26 & 77.69 ± 0.17 \\
Prototypes + TRAML~\citep{li2020boosting} & ResNet-12 & 60.31 ± 0.48 & 77.94 ± 0.57 \\

\rowcolor{myblue}
\texttt{HyperProto} (Ours) & ResNet-12 & 59.65 ± 0.62 & 78.24 ± 0.47 \\
\midrule

ProtoNet~\citep{snell2017prototypical} & WideResNet-28-10  & 59.09 ± 0.64  &  79.09 ± 0.46   \\ 
Activation to Parameter~\citep{qiao2018few} & WideResNet-28-10 & 59.60 ± 0.41 & 73.74 ± 0.19  \\
LEO ~\citep{rusu2018meta} & WideResNet-28-10 & 61.76 ± 0.08 & 77.59 ± 0.12  \\
SimpleShot~\citep{wang2019simpleshot} & WideResNet-28-10  & {63.50 ± 0.20} &   {80.33 ± 0.14}   \\
AWGIM~\citep{guo2020attentive} & WideResNet-28-10  & 63.12 ± 0.08  &  78.40 ± 0.11 \\
\rowcolor{myblue}
\texttt{HyperProto} (Ours) & WideResNet-28-10  & \textbf{63.78 ± 0.63} &   \textbf{81.29 ± 0.46}   \\
\bottomrule

\end{tabular}}
\caption{Accuracies with 95\% confidence interval on 1000 test episodes of \texttt{HyperProto} and baselines on \textit{mini}ImageNet.}
\label{tab:image}
\end{table*}

\subsection{Comparison with Other Approaches}
\label{subsec:ner}
\looseness=-1 \textbf{Few-shot Named Entity Recognition.} Table \ref{tab:fewnerd} shows the performance of other methods on \textsc{Few-NERD}. Overall, \texttt{HyperProto} has a considerable advantage over vanilla ProtoNet, with an increase of at least 5\% in f1-score across all settings. The success on both datasets demonstrates that \texttt{HyperProto} can learn the general distribution pattern of entities across different classes and thus can greatly improve the performance when little information is shared between the training and test set. It can also be observed that a large portion of the improvement comes from the increase in precision, indicating the ability of \texttt{HyperProto} to distinguish entities from context. It is possibly because context words are very diverse, and modeling them with a hypersphere as prototypes is more fitting than a single point as in ProtoNet. With respect to the number of shots, \texttt{HyperProto} is more advantageous when larger shots are provided and becomes the new state-of-art in the 5$\sim$10 shot setting. For the comparison with NNShot, \texttt{HyperProto} remains superior under the settings of high-shot (5$\sim$10), outperforming it by at least 10\% of the F1 score. 
Interestingly, the performances of NNShot and \texttt{HyperProto} are comparable when it comes to low-shot. This is probably because, in the sequence labeling task, it is more difficult to infer the class-level information from very limited tokens. 
In this case, the modeling ability of hypersphere prototypes degenerates towards the nearest-neighbors strategy in NNShot.
As the shot number increases, the memory cost of NNShot grows quadratically and becomes unaffordable, while \texttt{HyperProto} keeps it in reasonable magnitude. In this sense, \texttt{HyperProto} is more  efficient. 
We also believe a carefully designed initialization strategy is vital for the performance of our model in low-shot settings. The impact of the number of shots is reported in Appendix \ref{appendix:shot}.

\looseness=-1 \noindent \textbf{Few-shot Relation Extraction.} Table \ref{tab:fewrel1.0} presents the results on two FewRel tasks. Methods that use additional data or conduct task-specific encoder pre-training are not included. 
\texttt{HyperProto} generally performs better than all baselines across all settings. In terms of backbone models, when combined with pre-trained models like BERT, hypersphere prototypes can yield a larger advantage against prototypes. It shows that the  hypersphere modeling of prototypes can better approximate the real data distribution and boosts the finetuning of BERT. 
Meanwhile, it sheds light on the untapped ability of large pre-trained language models and stresses that a proper assumption about data distribution may help us unlock the potential. 
\texttt{HyperProto}'s outstanding performance on the Domain Adaptation task further validates the importance of a better abstraction of data in transfer learning. Meanwhile, the large performance variation in the domain adaptation task suggests that when the domain shifts, the estimation of hypersphere prototypes becomes less stable.

To further evaluate the compatibility of our approach and other orthogonal techniques, we conduct task-specific pre-training for relation extraction, where 1,000,000 instances are collected as the training data in a distantly supervised manner from Wikipedia. In the data-collecting process, we ensure that no instances from FewRel and FewRel 2.0 are involved. 
Then we directly train a BERT encoder with the naive cross-entropy objective function to obtain the final encoder. It could be observed that, with this pre-trained encoder, the performance of our method boosts substantially, demonstrating the model-agnostic nature of our approach.


                        


\looseness=-1 \noindent \textbf{Few-shot Image Classification.} Table~\ref{tab:image} shows the result on \textit{mini}ImageNet few-shot classification under 2 settings. \texttt{HyperProto} substantially outperforms the primary baselines in most settings, displaying their ability to model the class distribution of images. We observe that compared to NLP, image classification results are more stable both for vanilla prototypes and hypersphere prototypes. This observation may indicate the difference in encoding between the two technologies. Token representations in BERT are contextualized and changeable around different contexts, yet the image representation produced by deep CNNs aims to capture the global and local features thoroughly.
Under the 5-way 5-shot setting, the improvements of \texttt{HyperProto} are significant. 
The effectiveness of our method is also demonstrated by the comparisons with other previous few-shot learning methods with the same backbones. In particular, \texttt{HyperProto} yields the best results of all the compared methods with the WideResNet~\citep{Zagoruyko2016WRN} backbone, suggesting that 
the expressive capability of hypersphere prototypes can be enhanced with a more powerful encoder. Compared to the 5-shot setting, our model improves mediocrely
in the 1-shot setting of ConvNet and ResNet-12~\citep{he2015deep}. The phenomenon is consistent with the intuition that more examples would be more favorable to the learning of radius.
We further analyze the dynamics of the radius of our method in Appendix~\ref{appendix:radius}.

\subsection{Experimental Analysis}



\begin{table}[h]
\centering

\scalebox{0.85}{
\begin{tabular}{lcc} \toprule
\multirow{2}{*}{\textbf{ {Methods}}}  & \multicolumn{2}{c}{\textbf{\textit{mini}ImageNet}}                                               \\ \cmidrule(r){2-3}
                     & \multicolumn{1}{c}{ {\textbf{5 way 1 shot}}}  & \multicolumn{1}{c}{ {\textbf{5 way 5 shot}}}  \\
                    \midrule  

\rowcolor{myblue}
Cone \texttt{HyperProto}   & 62.43 ± 0.63 & 76.03 ± 0.50 \\
\rowcolor{myblue}
Gaussian \texttt{HyperProto}   & 60.34 ± 0.64  &  80.43 ± 0.45   \\
\rowcolor{myblue}
\texttt{HyperProto}   & \textbf{63.78 ± 0.63} &   \textbf{81.29 ± 0.46}   \\
\bottomrule
\end{tabular}}

\caption{\looseness=-1 Accuracies with 95\% confidence interval of generalized \texttt{HyperProto} on \textit{mini}ImageNet.}

\label{tab:general_exp}
\end{table}

\noindent \looseness =-1 \textbf{Comparison of Generalized Prototypes.} To further compare the variants of our approach, we conduct experiments for cone-like and gaussian prototypes with WideResNet-28-10 on \textit{mini}ImageNet as well.
Table~\ref{tab:general_exp} presents results across three measurement settings. Although the two variants do not perform better than our main method, they still considerably outperform many baselines in Table~\ref{tab:image}. 
While the three models' performance is close under the 1-shot setting, the cone-like prototypes model performs worse in the 5-shot setting. It could be attributed to unsatisfying radius learning. It is found that the cone-like prototypes model is susceptible to radius learning rate and is prone to overfitting. 

\begin{table}[h]
    \centering

    \scalebox{0.74}{
    \begin{tabular}{lcc}
    \toprule
    \multirow{2}{*}{\textbf{Methods}} & \textbf{Backbone} & \textbf{5 way 5 shot}  \\
    \cmidrule{2-3}
     & \multicolumn{2}{c}{\textbf{\textit{mini}ImageNet} $\rightarrow$ \textbf{CUB}} \\ \midrule
    MatchingNet~\citep{vinyals2016matching} & RN-12& 53.07 ± 0.74 \\
    ProtoNet~\citep{snell2017prototypical} &RN-12&  62.02 ± 0.70 \\
    MAML~\citep{finn2017model} & RN-18& 52.34 ± 0.72 \\
    RelationNet~\citep{sung2018learning} &RN-18&  57.71 ± 0.73 \\
    Baseline++~\citep{chen2021meta} & RN-18 & 62.04 ± 0.76 \\
    SimpleShot~\citep{wang2019simpleshot} & RN-18 & 65.56 ± 0.70 \\
    \midrule
    \rowcolor{myblue}
    \texttt{HyperProto} (Ours) & RN-12&   63.22 ± 0.77 \\
    \bottomrule
    \end{tabular}}
    \caption{\looseness=-1 Results on cross-dataset classification. }
    \label{tab:exp-cross}
\end{table}

\noindent \textbf{Cross-dataset Few-shot Learning.}  We also conduct experiments on the more difficult cross-dataset setting. Specifically, the model trained on \textit{mini}Imagenet is tested on the CUB dataset~\citep{welinder2010caltech} under the 5-way 5-shot setting. We use ResNet-12 (RN-12)~\citep{he2015deep} as the backbone in our experiment.
Table~\ref{tab:exp-cross} shows the results compared with several baselines. It can be seen that \texttt{HyperProto} outperforms the baselines by a large margin even with less powerful encoder (RN-12), indicating the ability to learn representations that are transferrable to new domains. The results also echo the performance of \texttt{HyperProto} for the cross-domain relation extraction in Table~\ref{tab:fewrel1.0}.

\begin{figure}[h]
\centering
\includegraphics[width=0.98\linewidth]{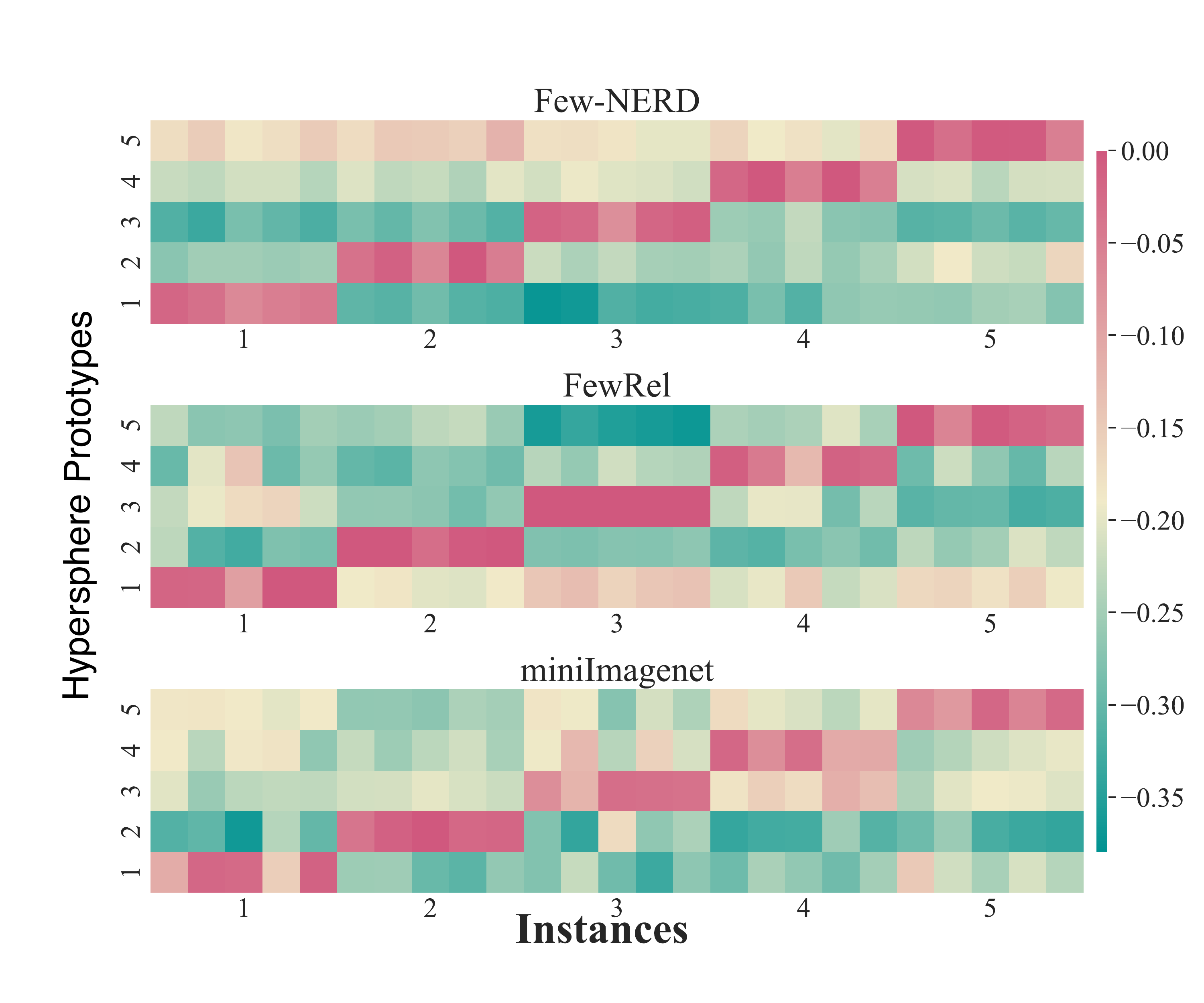}
\caption{\looseness=-1 Normalized distances from instances to hypersphere prototypes. Horizontal axis: hypersphere prototypes of 5 classes. Vertical axis: 5 instances per class.}
\label{fig:heat}
\end{figure}

\begin{figure*}[!th]
    \centering
    \includegraphics[width=1\textwidth]{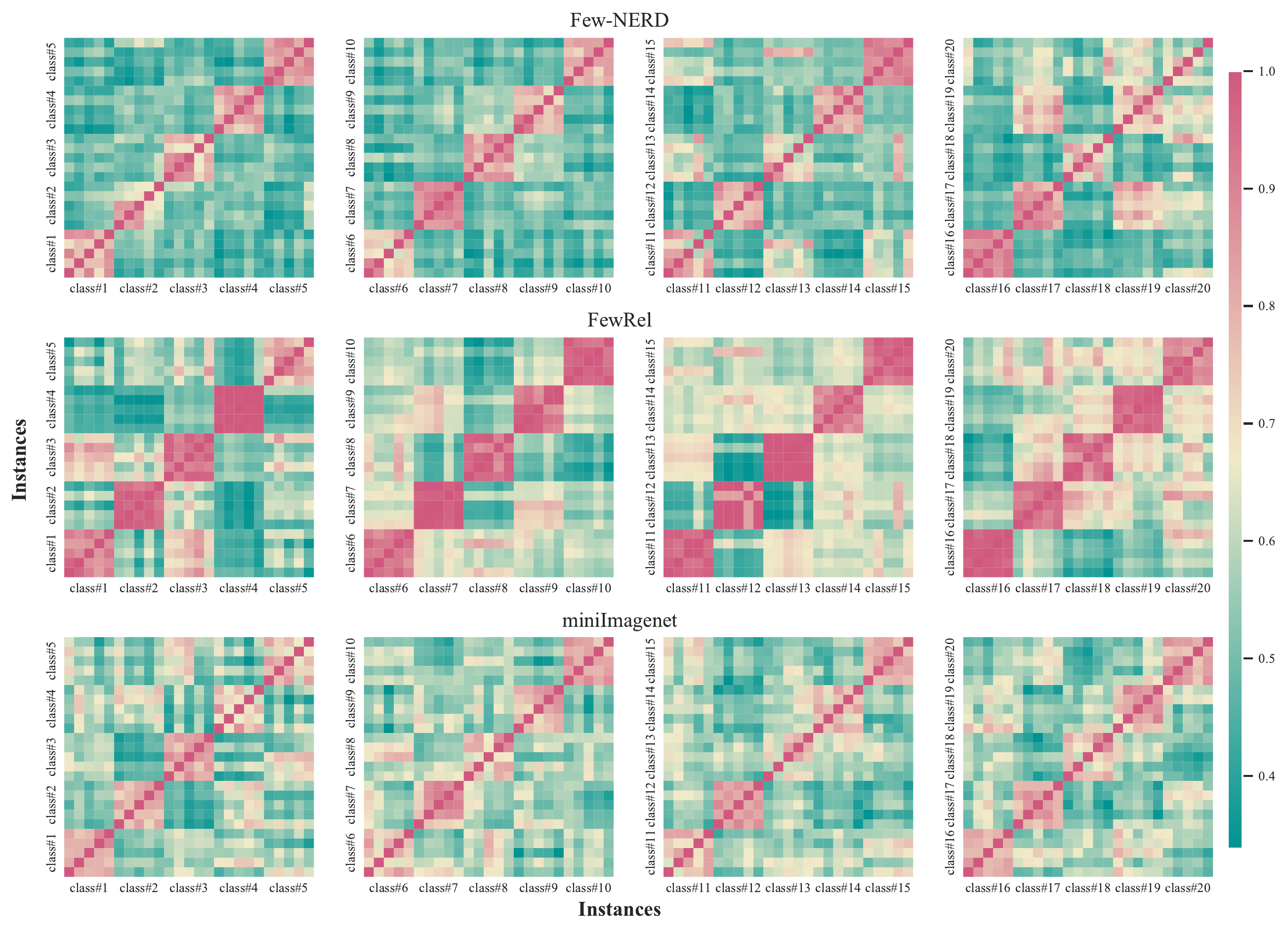}
    \caption{Representation similarity matrix produced by \texttt{HyperProto} on \textsc{Few-NERD}, FewRel and \textit{mini}ImageNet. Each row illustrates 20 classes and 100 instances in one dataset. Each subfigure contains 5 classes and 25 instances. 
    Each unit denotes the cosine similarity of two embeddings, and each 5$\times$5 cell indicates the comparison of two classes.  The units on the diagonal represent the same instance, and the 5$\times$5 cells on the diagonal represent the same class.
    Warmer color means higher similarity in this illustration. }
    \label{fig:instance}
\end{figure*}

\looseness=-1 \noindent \textbf{Representation Analysis.} To study if the learned representations are discriminative, we illustrate the normalized distances between the learned representations and the hypersphere prototypes in Figure~\ref{fig:heat}. Specifically, we randomly sample 5 classes and 25 instances (5 per class) for each dataset and produce representations for the instances and hypersphere prototypes for the classes. Then, we calculate the distance between each instance to each prototype (i.e., distance from the point to the hypersphere surface) to produce the matrix. All the values in the illustration are normalized since the absolute values may 
vary with the datasets. Warmer colors denote shorter distances in the illustration. 
The illustration shows that in all three datasets, our model could effectively learn discriminative representations and achieve stable metric-based classification.
In order to further analyze the representations produced by \texttt{HyperProto}, we study the similarities of randomly sampled instance embeddings. We randomly select 4 $\times$ 5 classes and 5 instances per class in \textsc{Few-NERD}, FewRel and \textit{mini}ImageNet, respectively. As illustrated in Figure~\ref{fig:instance}, each subfigure is a 25 $\times$ 25 matrix based on 5 classes. We calculate the cosine similarities of these embeddings and observe clear intra-class similarity and inter-class distinctiveness. This result confirms the robustness of our model since all the classes and instances are sampled randomly.

\subsection{Analysis of the Radius Dynamics}
\label{appendix:radius}

In this section, the mechanism of hypersphere prototypes will be empirically analyzed. We demonstrate the mechanism of hypersphere prototypes by illustrating the change of radius for one specific hypersphere. In the learning phase, the radius of a hypersphere prototype is changing according to the ``density'' of the sampled episode, which could be characterized by the mean distance of samples to the corresponding prototype center. Practically, due to randomness in sampling, the value of the mean distance may oscillate at a high frequency in this process, and the radius changes accordingly. 
To better visualize the changing of radius along with the mean distance at each update, for each round of training, we fix one specific class as the \textit{anchor class} for mean distance and radius recording and apply a special sampling strategy at each episode. Specifically, we take FewRel training data and train on the 5 way 5 shot setting with CNN encoder. While training, each episode contains the \textit{anchor class} and 4 other randomly sampled classes. Training accuracy is logged every 50 steps. After a warmup training of 500 steps, we sample ``good'' or ``bad'' episodes for every 50 steps alternatively. A ``good'' episode has higher accuracy on the anchor class than the previously logged accuracy, while conversely, a ``bad'' episode has an accuracy lower than before. The mean distance to the prototype center and radius at each episode are logged every 50 steps after the warmup.
\begin{figure*}[!th]
 	\centering
 		\includegraphics[width=1.0\linewidth]{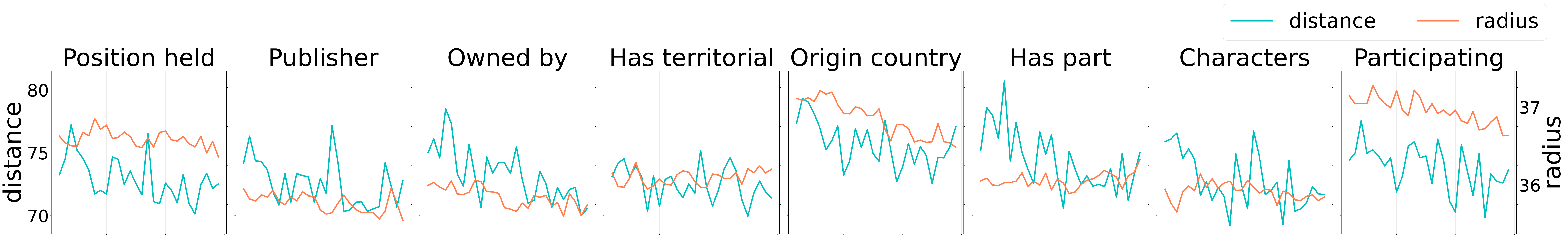}
 		\caption{The illustration depicts the radius change according to the degree of sparsity of the sampled episode. Each subfigure represents a selected anchor class in the FewRel dataset. The horizontal axis represents the increase of training steps.}
\label{fig:radius}
\end{figure*}
Figure~\ref{fig:radius} shows the changing of mean distance and radius for 8 classes during 600$\sim$2000 training steps. Although the numeric values of distance and radius differ greatly and oscillate at different scales, they have similar changing patterns. Besides, it could be observed that there is often a small time lag in the change of radius, indicating that the change of radius is brought by the change in mean distance. This is in line with our expectations and perfectly demonstrates the learning mechanism of hypersphere prototypes. The experiment provides a promising idea, if we can control the sampling strategy through knowledge a priori, we may find a way to learn ideal hypersphere prototypes.


\section{Related Work}
\label{sec:rw}
This work is related to studies of meta-learning, whose primary goal is to quickly adapt deep neural models to new tasks with a few training examples~\citep{hospedales2020meta}. 
To this end, two branches of studies are proposed: optimization-based methods and metric-based methods. The optimization-based studies~\citep{finn2017model, franceschi2018bilevel, ravi2018amortized} regard few-shot learning as a bi-level optimization process, where a global optimization is conducted to learn a good initialization across various tasks, and a local optimization quickly adapts the initialization parameters to specific tasks by a few steps of gradient descent. 

\looseness=-1 Compared to the mentioned studies, our work is more related to the metric-based meta-learning approaches~\citep{vinyals2016matching, snell2017prototypical, satorras2018few, sung2018learning}, whose general idea is to learn to measure the similarity between representations and find the closest labeled example (or a derived prototype) for an unlabeled example. Typically, these methods learn a measurement function during episodic optimization.
More specifically, we inherit the spirit of using prototypes to abstractly represent class-level information, which could be traced back to cognitive science~\citep{reed1972pattern, rosch1976basic, nosofsky1986attention}, statistical machine learning~\citep{graf2009prototype} and to the Nearest Mean Classifier~\citep{mensink2013distance}. In the area of deep learning, \citet{snell2017prototypical} propose the prototypical network to exploit the average of example embeddings as a prototype to perform metric-based classification in few-shot learning. In their work, prototypes are estimated by the embeddings of instances. However, it is difficult to find a satisfying location for the prototypes based on the entire dataset. \citet{ren2018meta} adapt such prototype-based networks in the semi-supervised scenario where the dataset is partially annotated. Moreover, a set of prototype-based networks are proposed concentrating on the improvements of prototype estimations and application to various downstream tasks~\citep{allen2019infinite, gao2019hybrid, li2019prototype, pan2019transferrable, seth2019prototypical, ding2020prototypical, li2020prototypical, wertheimer2019few, xie2022joint, zhang2020deepemd}. We discuss our method within the context of other prototype-enhanced methods in Appendix~\ref{appendix:other}. There has also been a growing body of work that considers the few-shot problem from multiple perspectives, bringing new thinking to the field~\citep{tian2020rethink,  yang2020free, laenen2021episodes, zhang2020iept, wang2021grad2task, das2021container, wertheimer2021few, ding2021prompt, Cui2022PrototypicalVF, Hu2022KnowledgeablePI}.

\looseness=-1 There has also been a series of works that embed prototypes into a non-Euclidean output space~\citep{mettes2019hyperspherical, keller2020theory, atigh2021hyperbolic}. 
It should be noted that these studies regard hyperspheres or other non-Euclidean manifolds as a characterization of the embedding space, while our proposed method use hyperspheres to represent prototypes and conduct metric-based classification in the Euclidean space. Therefore, the focus of our proposed  \texttt{HyperProto} is different from the above non-Euclidean prototype-based works.

\section{Conclusion}
\label{sec:conclusion}

\looseness=-1 This paper proposes a novel metric-based few-shot learning method, \textit{hypersphere prototypes}. Unlike previous metric-based methods that use dense vectors to represent the class-level semantics, we use hyperspheres to enhance the capabilities of prototypes to express the intrinsic information of the data. It is simple to model a hypersphere in the embedding space and conduct metric-based classification in few-shot learning. Our approach is easy to implement and also empirically effective, we observe significant improvements to baselines on three tasks across NLP and CV. We also develope two variants of prototypes in other embedding spaces. For potential future work, such modeling could be extended to more generalized representation learning like word embeddings.

\bibliography{anthology,custom}
\bibliographystyle{acl_natbib}

\clearpage
\appendix


\section{Additional Experiments and Analysis}
\label{appendix:additional_exp}
This section provides additional experiments and analysis, we first visualize and quantify the representations learned by our approach. Then we analyze the dynamics of the radius parameter during training. At last, we conduct representation analysis at the instance level. 

\subsection{Visualization.} 
\begin{figure}[!th]
\centering
\includegraphics[width = 0.9\linewidth]{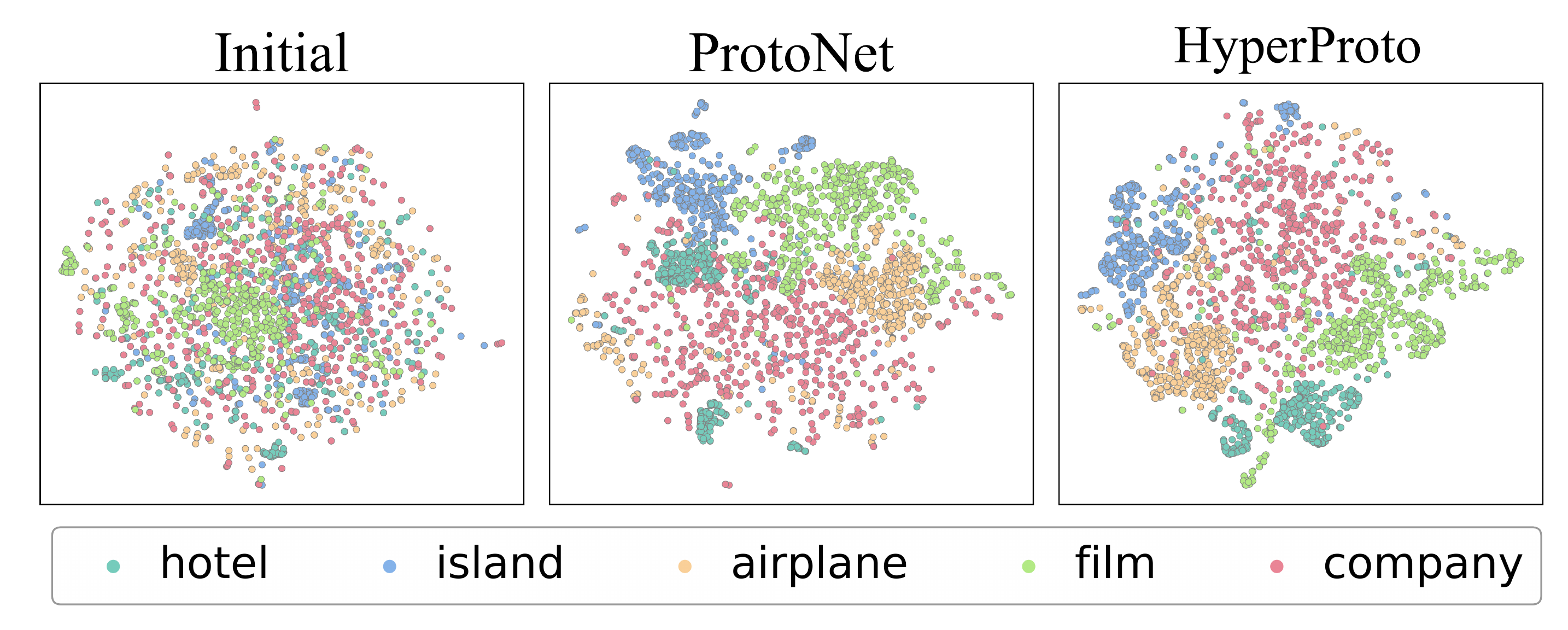}%
\hspace{0.15cm}%
\hspace{0.1cm}%
\includegraphics[width = 0.9\linewidth]{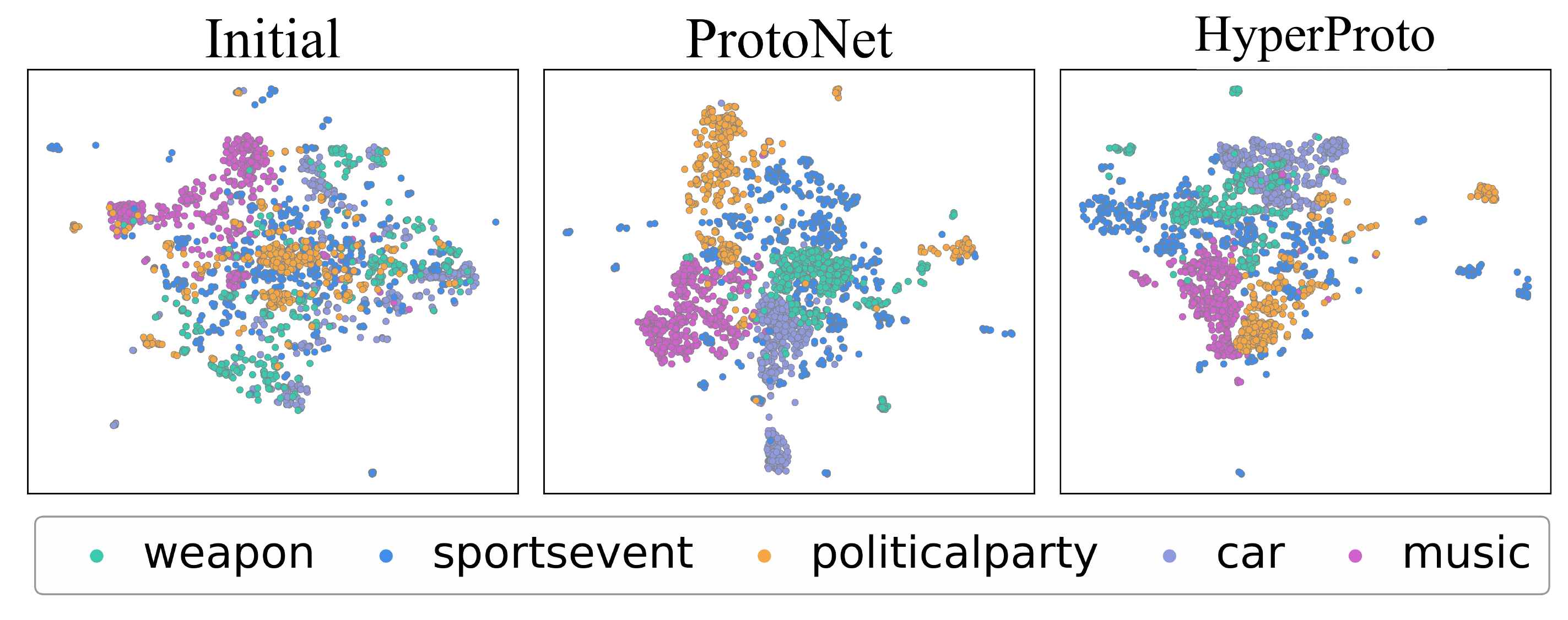}
\caption{$t$-sne visualization of feature distributions. The six subfigures, from left to right, are the representations of seen data (in training set) before training, produced by ProtoNet, and produced by \texttt{HyperProto}; novel data (in test set) before training, produced by ProtoNet, and produced by \texttt{HyperProto}. Note that even after training, the neural network has never seen the novel data and their classes. }
\label{fig:tsne}
\end{figure}

We also use $t$-SNE~\citep{vandermaaten2008visualizing} to visualize the embedding before and after training, by ProtoNet and \texttt{HyperProto}, respectively. 5 classes are sampled from the training set and test set of the Few-NERD dataset, and for each class, 500 samples are randomly chosen to be embedded by BERT trained on the 5-way-5-shot NER task. Figure~\ref{fig:tsne} shows the result of embeddings in a 2-dimensional space, where different colors represent classes. 
Note that for the token-level NER task, the interaction between the target token and its context may result in a more mixed-up distribution compared to instance-level embedding. 
For both models, the representations of the same class in the training set become more compressed and easier to classify compared to their initial embeddings. While \texttt{HyperProto} can produce even more compact clusters. The clustering effect is also observed for novel classes. 
We also calculate the difference between the mean euclidean distances from each class sample to the (hypersphere) prototype of the target class and to other classes. The larger the difference, the better the samples are distinguished. For ProtoNet, the difference is 2.33 and 1.55 on the train and test set, while for \texttt{HyperProto} the results are 5.09 and 4.56, respectively. This can also be inferred from the \textit{t}-sne result. Since samples from different classes are distributed at different densities, an extra radius parameter will help better distinguish between classes.
The visualization and statistical results demonstrate the effectiveness of \texttt{HyperProto} in learning discriminative features, especially in learning novel class representation that considerably boosts model performance under few-shot settings.

\subsection{Impact of Number of Shots}
\label{appendix:shot}
\begin{figure}[!ht]
    \centering
    \includegraphics[width=0.45\textwidth]{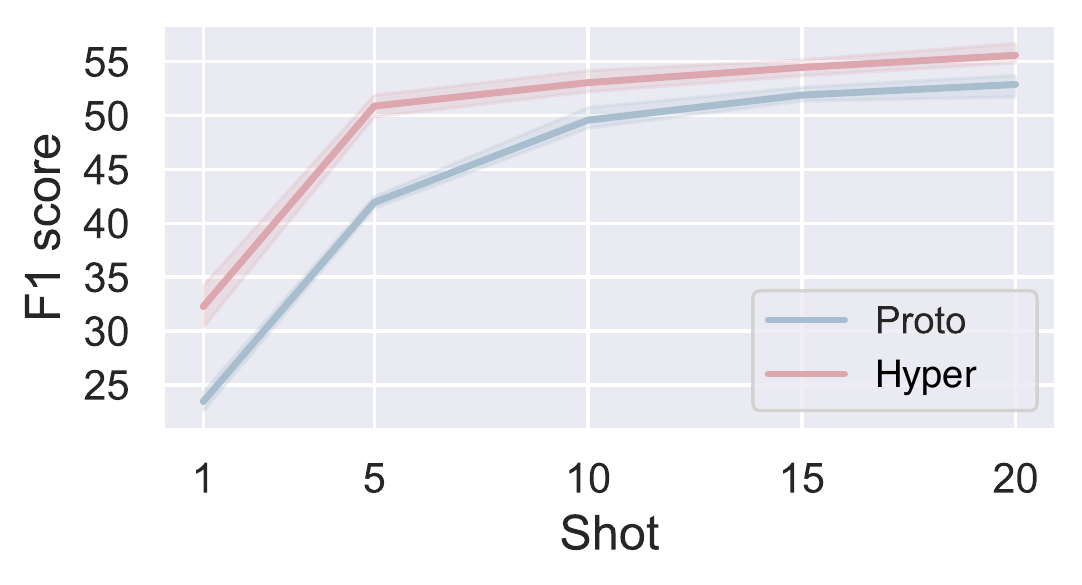}
    \caption{Impact of shot number on model performance for \textsc{Few-NERD (INTRA)} 5-way setting.}
    \label{fig:shot}
\end{figure}

We conduct additional experiments on \textsc{Few-NERD (INTRA)} 5-way setting with 10, 15, 20 shots. Since NNShot becomes too memory-intensive to run when shot reaches 15, we provide results on Proto and \texttt{HyperProto}. Figure~\ref{fig:shot} shows both models perform better when more data are available, while \texttt{HyperProto} performs consistently better than vanilla prototypes.

\section{Experimental Details}
\label{appendx:exp}

This section reports the experimental details of all three tasks in our evaluation. All the experiments are conducted on NVIDIA A100 and V100 GPUs with CUDA. The main experiments are conducted on three representative tasks in NLP and CV, which are few-shot named entity recognition (NER), relation extraction (RE), and image classification. The experimental details will be presented in the following sections.


\subsection{Experimental Details for Few-shot Named Entity Recognition}
\label{appendx:exp_ner}

\looseness=-1 We assess the effectiveness of hypersphere prototypes on NLP, specifically, the first task is few-shot named entity recognition (NER) and the dataset is \textsc{Few-NERD}~\citep{ding2021fewnerd}\footnote{\textsc{Few-NERD} is distributed under CC BY-SA 4.0 license}. 
NER aims at locating and classifying named entities ( real-world objects that can be denoted with proper names) given an input sentence, which is typically regarded as a sequence labeling task. Given an input sentence "``Bill Gates is a co-founder of the American multinational technology corporation {Microsoft}'', an named entity recognition system aims to locate the named entities (\textit{Bill Gates}, \textit{Microsoft}) and classify them into specific types. Conventional schema uses coarse-grained labels such that \textit{Person} for \textit{Bill Gates} and \textit{Organization} for \textit{Microsoft}. In more advanced schema like Few-NERD, models are asked to give more specific entity types, for example, \textit{Person-Entrepreneur} for \textit{Bill Gates} and \textit{Organization-Company} for \textit{Microsoft}. 
Different from typical instance-level classification, few-shot NER is a sequence labeling task, where labels may share structural correlations. NER is the first step in automatic information extraction and the construction of large-scale knowledge graphs. Quickly detecting fine-grained unseen entity types is of significant importance in NLP.
To capture the latent correlation, many recent efforts in this field use large pre-trained language models~\citep{han2021pre} like BERT~\citep{devlin2018bert} as backbone model and have achieved remarkable performance. The original prototypical network has also been applied to this task~\citep{li2020few, huang2020few, de2021meta}. 

\looseness=-1 \noindent \textbf{Dataset.} The experiment is run on \textsc{Few-NERD} dataset~\citep{ding2021fewnerd}. It is a large-scale NER dataset containing over 400,000 entity mentions, across 8 coarse-grained types and 66 fine-grained types, which makes it an ideal dataset for few-shot learning. It has been shown that existing methods including prototypes are not effective enough on this dataset.

\looseness=-1 \noindent \textbf{Baselines.} NNShot~\citep{yang-katiyar-2020-simple} is a token-level metric-based method that is specifically designed for few-shot labeling. Note that the main baseline here is the Proto method, which adapts the prototypical network on few-shot named entity recognition.

\looseness=-1 \noindent \textbf{Implementation Details.} We run experiments under four settings on the two released benchmarks, \textsc{Few-NERD (INTRA)} and \textsc{Few-NERD (INTER)}. Specifically, we use uncased BERT as the backbone encoder and 1e-4 as the encoder learning rate. We manually tune the learning rate for the radius parameter, and the best result is obtained with 10. AdamW is used as the BERT optimizer, and Adam~\citep{kingma2017adam} is used to optimize prototype radius. The batch size is set to 2 across all settings. All models are trained for 10000 steps and validated every 1000 steps. The results are reported on 5000 steps of the test episode. For each setting, we run the experiment with 3 different random seeds and report the average results including precision, recall, f1-score, and the standard error for each. We use PyTorch~\citep{paszke2019pytorch} and huggingface transformers~\citep{wolf2019huggingface} to implement the backbone encoder $\text{BERT}_{\text{base}}$.

\subsection{Experimental Details for Few-shot Relation Extraction}
\label{appendx:exp_re}

\looseness=-1 The other common NLP task is relation extraction (RE), which aims at correctly classifying the relation between two given entities in a sentence. For example, given an input sentence with marked entities ``[\textit{Bill Gates}] is a co-founder of the American multinational technology corporation [\textit{Microsoft}]'', the relation extraction system aims to give the relationship between \textit{Bill Gates} and \textit{Microsoft}. 
This is a fundamental task in information extraction. RE is an important form of learning structured knowledge from unstructured text.  We use FewRel~\cite{han2018fewrel}\footnote{FewRel is distributed under MIT license} and FewRel 2.0~\cite{gao2019fewrel} as the datasets.
In real-world datasets, many of the relations are long-tailed and thus cannot be identified accurately under the common supervised setting. Traditional methods often alleviate the problem with distant supervision, which would result in wrong labels. Recent approaches have applied few-shot learning models on the task to learn from a handful of samples, which yield promising results~\citep{gao2019hybrid}. We report the datasets, baselines, and experimental details in Appendix~\ref{appendx:exp_re}.

\looseness=-1 \noindent \textbf{Dataset.}
We adopt the FewRel dataset \citep{han2018fewrel, gao2019fewrel}, a relation extraction dataset specifically designed for few-shot learning. FewRel has 100 relations with 700 labeled instances each. The sentences are extracted from Wikipedia and the relational entities are obtained from Wikidata. FewRel 1.0~\citep{han2018fewrel} is released as a standard few-shot learning benchmark. FewRel 2.0~\citep{gao2019fewrel} adds domain adaptation task and NOTA task on top of FewRel 1.0 with the newly released test dataset on PubMed corpus.

\looseness=-1 \noindent \textbf{Baselines.} In addition to the main baseline, prototypical network~\cite{snell2017prototypical}, we also choose the following few-shot learning methods as the baselines in relation extraction. (1) Proto-HATT~\cite{gao2019hybrid} is a neural model with hybrid prototypical attention. (2) MLMAN~\cite{ye2019multi} is a multi-level matching and aggregation network for few-shot relation classification. Note that Proto-HATT and MLMAN are not model-agnostic.
(3) GNN~\cite{satorras2018few} is a meta-learning model with a graph neural network as the prediction head. (4) SNAIL~\cite{mishra2017simple} is a meta-learning model with attention mechanisms. (5) Meta Net~\cite{munkhdalai2017meta} is a classical meta-learning model with meta information. (6) Proto-ADV~\cite{gao2019fewrel} is a prototype-based method enhanced by adversarial learning. (7) BERT-pair~\cite{gao2019fewrel} is a strong baseline that uses BERT for few-shot relation classification. We re-run all the baselines, except for MLMAN, and report the corresponding performances.

\noindent \textbf{Implementation Details} The experiments are conducted on FewRel 1.0 and FewRel 2.0 domain adaptation tasks. For FewRel 1.0, we follow the official splits in~\citet{han2018fewrel}. For FewRel2.0, we follow~\citet{gao2019fewrel}, training the model on wiki data, validating on SemEval data, and testing on the PubMed data. We use the same CNN structure and BERT as encoders. The learning rate for hypersphere prototype radius is 0.1 and 0.01 for CNN and BERT encoder, respectively. Adam~\citep{kingma2017adam} is used as radius optimizer. We train the model for 10000 steps, validate every 1000 steps, and test for 5000 steps. The other hyperparameters are the same as in the original paper. 

\subsection{Experimental Details for Few-shot Image Classification}
\label{appendx:exp_ic}

Image classification is one of the most classical tasks in few-shot learning research. Seeking a better solution for few-shot image classification is beneficial in two ways: (1) to alleviate the need for data augmentation, which is a standard technique to enrich the labeled data by performing transformations on a given image; (2) to facilitate the application where the labeled image is expensive. We use \textit{mini}ImageNet~\citep{vinyals2016matching} as the dataset in our experiment.
The dataset, baselines and experimental details are reported in Appendix~\ref{appendx:exp_ic}.

\noindent \textbf{Dataset.}
\textit{mini}ImageNet~\citep{vinyals2016matching} is used as a common benchmark for few-shot learning. The dataset is extracted from the full ImageNet dataset~\citep{Deng2009imagenet}, and consists of 100 randomly chosen classes, with 600 instances each. Each image is of size 3$\times$84$\times$84. We follow the split in ~\citep{Sachin2017} and use 64, 16, and 20 classes for training, validation, and testing. 

\noindent \textbf{Baselines.} The baselines we choose are as follows: (1) Prototypical network~\citep{snell2017prototypical} is our main baseline; (2) IMP~\citep{allen2019infinite} is a prototype-enhanced method that models an infinite mixture of prototypes for few-shot learning; (3) CovaMNet~\citep{li2019CovaMNet} is a few-shot learning method that uses covariance to model the distribution information to enhance few-shot learning performance.
(4) SNAIL~\citep{mishra2017simple} is an attention-based classical meta-learning method; (5) Variational FSL~\citep{zhang2019variational} is a variational Bayesian framework for few-shot learning, which contains a pre-training stage; (6) Activation to Parameter~\citep{qiao2018few} predicts parameters from activations in few-shot learning; (7) LEO~\citep{rusu2018meta} optimizes latent embeddings for few-shot learning. (8) TRAML~\citep{li2020boosting} uses adaptive margin loss to boost few-shot learning, and Prototypes + TRAML is a strong baseline in recent years.;
(9) Meta-baseline~\cite{chen2021meta} is a pre-training \& tuning method that serves as a strong baseline in few-shot learning.

\noindent \textbf{Implementation Details.} 
The experiments are conducted on 5 way 1 shot and 5 way 5 shot settings.  
To ensure validity and fairness, we implement hypersphere prototypes with various backbone models including CNN, ResNet-12, and WideResNet~\citep{Zagoruyko2016WRN} to make it comparable to all baseline results, and we also re-run some of the baselines including prototypical network~\citep{snell2017prototypical}, infinite mixture prototypes~\citep{allen2019infinite}, and CovaMNet~\citep{li2019CovaMNet} under our settings based on their original code. Other baseline results are taken from the original paper. Each model is trained on 10,000 randomly sampled episodes for 30$\sim$40 epochs and tested on 1000 episodes. The result is reported with 95\% confidence interval. Note that both ResNet-12 and WideResNet~\citep{Zagoruyko2016WRN} are pretrained on the training data, where the pretrained ResNet-12 is identical to~\citet{chen2021meta} and the pretrained WideResNet follows~\citet{mangla2020charting}. The CNN structure is the same as~\citet{snell2017prototypical}, which is composed of 4 convolutional blocks each with a 64-filter 3 × 3 convolution, a batch normalization layer~\citep{ioffe2015batch}, a ReLU nonlinearity, and a 2 × 2 max-pooling layer. We use SGD optimizer for the encoder and Adam~\citep{kingma2017adam} optimizer for the prototype radius. The learning rate for the backbone model is 1e-3. The learning rate for radius is manually tuned and the reported result in Table~\ref{tab:image} has a learning rate of 10. For cone-like and gaussian prototypes, we use 1e-1 and 1e-3. At the training stage, the prototype center is re-initialized at each episode as the mean vector of the support embeddings. 

\section{Discussion}

This section discusses other related prototype-based methods in detail, as well as limitations and broader impact of our work.
\subsection{Other Prototype-enhanced Methods}
\label{appendix:other}
\looseness=-1 Here, we discuss the difference between hypersphere prototypes with four prototype-enhanced methods in few-shot learning: infinite mixture prototypes~\citep{allen2019infinite}, CovaMNet~\citep{li2019CovaMNet}, variational few-shot learning~\citep{zhang2019variational}, and two-stage ~\citep{Das2020ATA}. 

Infinite mixture prototypes~\citep{allen2019infinite} model each class as an indefinite number of clusters and the prediction is obtained by computing and comparing the distance to the nearest cluster in each class. This method is an intermediate model between prototypes and the nearest neighbor model, whereas hypersphere prototypes alleviate the over-generalization problem of vanilla prototypes with a single additional parameter that turns a single point modeling into a hypersphere. The essential prototype-based feature of hypersphere prototypes allows faster computation and easier training. 

CovaMNet~\citep{li2019CovaMNet} calculates local variance for each class based on support samples and conducts metric-based learning via covariance metric, which basically evaluates the cosine similarity between query samples and the eigenvectors of the local covariance matrix. To ensure the non-singularity of the covariance matrix, the feature of each sample is represented with a matrix, generated by a number of local descriptors, with each extracting a feature vector. Compared to hypersphere prototypes, both methods attempt to model more variance based on vanilla prototypes, while the idea of hypersphere prototypes is more straightforward and requires fewer parameters. On the other hand, the multi-channel features adopted by CovaMNet are less natural for NLP tasks.

Variational Few-Shot Learning~\citep{zhang2019variational} tackles the few-shot learning problem under a bayesian framework. In order to improve single point-based estimation, a class-specific latent variable representing the class distribution is introduced and is assumed to be Gaussian. The method parameterizes the mean and variance of the latent variable distribution with neural networks that take the feature of a single instance as input. The learning and inference processes are both conducted on the latent variable level. The method adopts variational inference and is built on modeling distribution as a latent variable, where the metric calculation highly relies on the Gaussian assumption. Hypersphere prototypes, on the other hand, model the distribution with a center vector and a radius parameter in the actual embedding space, which is more tangible and easier to calculate. It is worth noting that this work also points that a single embedding is insufficient to represent a class, and samples the prototype from a high-dimensional distribution. This is actually similar to our starting point, the difference is that our approach turns out to consider the problem from the geometric point of view based on the original embedding space, and proves that such simple geometric modeling could be very efficient in the few-shot scenarios. 

Two-Stage Approach first trains feature encoder and variance estimator on training data in an episodic manner with extracted absolute and relative features. Then in the second stage, training data are split into "novel" class, and base class, novel class prototypes are learned from both sample mean and base class features. The classification is carried out with integrated prototypes. This method improves on vanilla prototypes by extracting more features and combining information from base classes, but still follows single-point-based metric learning. Our approach extends a single point to a hypersphere in the embedding space and therefore, better captures within-class variance.

\begin{algorithm*}[!ht]
\caption{Greedy $N$-way $K$$\sim$$2K$-shot sampling algorithm for \textsc{Few-NERD}}
\label{alg:gre}
\KwIn{Dataset $\mathcal{X}$, Label set $\mathcal{Y}$, $N$, $K$}
\KwOut{output result}
$\mathcal{S}\leftarrow \varnothing$;  {\tcp{Init the support set}} 
{\tcp{Init the count of entity types}} 
\For{$i=1$ to $N$}{
  $\text{Count}[i] = 0$ \;
}
\Repeat{$\text{Count}_i \geq K$ for $i = 1$ to N }{
Randomly sample $(\bm{x}, \bm{y}) \in \mathcal{X}$ \; 

Compute $|\text{Count}|$ and $\text{Count}_i$ after update \;
    \eIf{$|\text{Count}| > N$ or $\exists \text{Count}[i] > 2K$}{
      Continue \;
    }{
      $\mathcal{S} = \mathcal{S} \bigcup (\bm{x}, \bm{y})$ \;
      Update $\text{Count}_i$ \;
    }
}
\end{algorithm*}

\subsection{Limitations}
\label{appendix:limitations}
Under the 1-shot setting, hypersphere prototypes will face challenges in estimating the radius in support sets, this is because the initial radius may be biased by the randomness of sampling. When the radius is set to exactly 0, the model will resemble a traditional prototypical network. In our empirical study, we find that setting radius could consistently yield more robust performance than traditional ways. Although not as large as the boost in the multi-shot setting, our method in the 1-shot scenario still delivers non-trivial results and exceeds most baselines (Table~\ref{tab:fewnerd}, Table~\ref{tab:fewrel1.0}, Table~\ref{tab:image}).

\subsection{Broader Impact}
\label{appendix:impact}
Our method focuses on the method of few-shot learning, which enables machine learning systems to learn with few examples, and could be applied to many downstream applications. The technique itself does not have a direct negative impact, i.e., its impact stems primarily from the intent of the user, and there may be potential pitfalls when the method is applied to certain malicious applications.

\section{K$\sim$2K Sampling for Few-NERD}
\label{appendix:sample}

In the sequence labeling task \textsc{Few-NERD}, the sampling strategy is slightly different from other classification tasks.
Because in the named entity recognition, each token in a sequence is asked to be labeled as if it is a part of a named entity. And the context is crucial for the classification of each entity, thus the examples are sampled at the sequence level. Under this circumstance, it is difficult to operate accurate $N$ way $N$ shot sampling. \citet{ding2021fewnerd} propose a greedy algorithm to conduct $N$ way $K\sim 2K$ shot sampling for the \textsc{Few-NERD} dataset.
We follow the strategy of the original paper~\cite{ding2021fewnerd} and report it in Algorithm~\ref{alg:gre}.

\end{document}